\documentclass[sn-mathphys,Numbered, iicol]{sn-jnl}

\usepackage{graphicx}%
\usepackage{multirow}%
\usepackage{amsmath,amssymb,amsfonts}%
\usepackage{amsthm}%
\usepackage{mathrsfs}%
\usepackage[title]{appendix}%
\usepackage{xcolor}%
\usepackage{textcomp}%
\usepackage{manyfoot}%
\usepackage{booktabs}%
\usepackage{algorithm}%
\usepackage{algorithmicx}%
\usepackage{algpseudocode}%
\usepackage{listings}%

\theoremstyle{thmstyleone}%
\theoremstyle{thmstyletwo}%

\theoremstyle{thmstylethree}%

\begin{document}

\title[Article Title]{A lightweight YOLOv5-FFM model for occlusion pedestrian detection}


\author[1]{\fnm{Xiangjie} \sur{Luo}}\email{luoxiangjie@buaa.edu.cn}

\author[1]{\fnm{Bo} \sur{Shao}}\email{ZB2103011@buaa.edu.cn}

\author*[1,2]{\fnm{Zhihao} \sur{Cai}}\email{czh@buaa.edu.cn}

\author[1,2]{\fnm{Yingxun} \sur{Wang}}\email{wangyx@buaa.edu.cn}

\affil[1]{\orgdiv{School of Automation Science and Electrical Engineering}, \orgname{Beihang University}, \orgaddress{\city{Beijing}, \postcode{100191}, \country{China}}}

\affil[2]{\orgdiv{Institute of Unmanned System}, \orgname{Beihang University}, \orgaddress{\city{Beijing}, \postcode{100191}, \country{China}}}


\abstract{The development of autonomous driving technology must be inseparable from pedestrian detection. Because of the fast speed of the vehicle, the accuracy and real-time performance of the pedestrian detection algorithm are very important.  YOLO, as an efficient and simple one-stage target detection method, is often used for pedestrian detection in various environments. However, this series of detectors face some challenges, such as excessive computation and undesirable detection rate when facing occluded pedestrians. In this paper, we propose an improved lightweight YOLOv5 model to deal with these problems. This model can achieve better pedestrian detection accuracy with fewer floating-point operations (FLOPs), especially for occluded targets. In order to achieve the above goals, we made improvements based on the YOLOv5 model framework and introduced Ghost module and SE block. Furthermore, we designed a local feature fusion module (FFM) to deal with occlusion in pedestrian detection. To verify the validity of our method, two datasets, Citypersons and CUHK Occlusion, were selected for the experiment. The experimental results show that, compared with the original yolov5s model, the average precision (AP) of our method is significantly improved, while the number of parameters is reduced by 27.9\% and FLOPs are reduced by 19.0\%.}




\maketitle

\section{Introduction}\label{sec1}

Pedestrian detection is a typical example of computer vision in daily life \citep{bib1}. In fact, pedestrian detection has direct applications in autonomous driving, crime detection, video surveillance and so on \citep{bib2}. The single-stage deep convolutional neural networks represented by YOLO show excellent performance in target detection  \citep{bib3, bib4, bib5} and significantly outperform previous hand-crafted methods and two-stage neural networks. For example, YOLO allows the deep convolutional neural networks (CNNs) \citep{bib6} to output a bounding box with any aspect ratio, regardless of the step size of the sliding window approach. Although two-stage detectors, represented by R-CNN, are capable of achieving state-of-the-art accuracy, they require more parameters to be calculated and take up more operating space, which severely weakens the real-time performance of the algorithm. 

YOLO \citep{bib7} detection architectures and Single Shot Multibox Detector (SSD) \citep{bib8} have been proposed for several years, which greatly reduce the amount of computation compared to R-CNN by eliminating the region suggestion step and regression both class and bounding boxes at the same time. The YOLO network uses pre-set default boxes of different scales to detect different scales of object instances. With this approach, YOLO can run smoothly without a high hardware configuration.

The YOLOv5 model is by far the best-known model in the YOLO family, and subsequent YOLO models have not made major changes to their backbone and neck networks. However, the YOLOv5 network still has its limitations. Firstly, the feature pyramid network (FPN) \citep{bib9} can deal with different scale targets effectively by upsampling, but it ignores the global information. Secondly, for mobile devices and embedded devices, the YOLOv5 model still requires most of the resources and computing power of the hardware, and it is difficult to run other algorithms or more complex deep-learning network models on this basis. Third, since YOLOv5 adopts a single-stage target detection method, it is easy for YOLOv5 to fail to identify targets with incomplete feature information, such as occlusion, which greatly affects the target detection accuracy. However, occlusion is a common situation when detecting pedestrians. Therefore, considering global information, lightweight the model, and dealing with occlusion are crucial for pedestrian detection tasks. 

Based on the above problems, we add SE block and Ghost module into YOLO network, optimize the loss function and design a new feature fusion module (FFM). More specifically, we investigate a lightweight occluded pedestrian detection network based on YOLOv5 architecture, which enjoys a higher accuracy than the YOLOv5 model while reducing the computation. The Squeeze-and-Excitation (SE) block can effectively consider the global information by weighting the channels \citep{bib10}. Ghost modules can generate more feature maps with fewer parameters, reduce the floating-point operations(FLOPs) while ensuring the accuracy \citep{bib11}. WIoU\underline{~}Loss is an attention-based bounding frame loss that can more accurately measure the similarity between the predicted box and the Ground Truth (GT) box, especially when multiple object parts are involved  \citep{bib12}. Considering that the head area of the pedestrian is not easy to be occluded in the actual environment, and combined with the leg area, all the position information of the pedestrian can be obtained, including the height and width of the pedestrian prediction box and the coordinates of the boundary points. Therefore, we used the improved YOLOv5 to detect the head and leg area of pedestrians, and we further propose a feature fusion module (FFM), which can restore the overall prediction boxes of pedestrians through the detected head regions and leg regions, and the module handles most occlusion problems well. Our work can be divided into the following three parts.

First, we propose a feature fusion module (FFM) based on feature matching, which detects the head and leg regions in the image respectively, and then fuse these features into multiple overall candidate boxes according to the inherent proportion of the human body, and the IoU of these candidate boxes are used to determine whether to repeat the detection of the same person.

Second, we replace all the Convolution modules in YOLOv5 neck network, including the convolution in the C3 module, with lightweight Ghost modules to achieve less floating-point operations(FLOPs) and maintain detection accuracy and speed.

Third, to utilize global information and weaken the impact of large and harmful gradients in extreme samples on the loss function, we added the SE block to YOLOv5 backbone network and redesigned the loss function.

Finally, we conducted comparative experiments on two widely used pedestrian detection datasets, and the results showed that our method greatly improves the detection accuracy while having fewer computational parameters and being lightweight. Experiments demonstrate the superiority of our proposed model.

\section{Related work}\label{sec2}

As a one-stage detector, YOLO \citep{bib7} removes the region suggestion step and is faster than two-stage detectors.  These single-stage methods use regression boxes of default size and proportion, which are more efficient, but the detection accuracy is affected to some extent. Recently, many papers have proposed a variety of improvements to the one-stage detection model. Some of these approaches aim to further simplify the model and reduce the calculation of parameters for easy use on mobile and embedded devices, while some others focus on how to improve the detection precision in more challenging environments.

In order to reduce the number of parameters and FLOPs, while maintaining competitive performance. MobileNetV1 \citep{bib13} replaces the traditional convolution with depth-wise convolution (DW-Conv), which can adopt a different convolution kernel for each input channel, such method greatly reduces model parameters and improves model inference speed and has become the main reference mode for subsequent light-weight CNNs \citep{bib14, bib15, bib16, bib17, bib18, bib19, bib20, bib21} and modern large CNNs \citep{bib22, bib23, bib24}. MobileNetv2 \citep{bib16} multiplexes image features, increases feature fusion, and proposes inverted residual block (IRB). ShuffleNet changes $1\times1$ convolutions to group convolutions \citep{bib25, bib26, bib27}, which leads to much less computation compared to MobileNet. However, this results in inadequate communication of information between the different groups.

For the task of real-time pedestrian detection, YOLO is a commonly used detection framework. \citep{bib28} is based on a multi-scale fusion strategy, designs a detection layer for small-scale targets, and integrates Resnet and Transformer structures. \citep{bib29} enables more accurate detection of overlapping targets by adjusting the IoU value of the Non-Maximum Suppression (NMS). \citep{bib30} combines the SE blocks into a new network structure, which pays good attention to the global information, the proposed model focuses more on the characteristics of small scale and occlusion. 

However, the above methods all directly use convolution to learn the overall features of pedestrians, which is also related to the currently commonly used datasets annotation methods, such as CityPersons \citep{bib31} and CUHK Occlusion Dataset \citep{bib32}, they all directly annotate the Ground Truth (GT) boxes of the overall pedestrians. Such a learning method has poor robustness and the overall features are too vulnerable. In the actual pedestrian detection process, pedestrians are easily occluded by various objects, and when the features are not obvious or incomplete, it will lead to false detection or missing detection. In our point of view, the final task of pedestrian detection is to detect the targets and return the overall prediction box of the target, including the obscured and unobscured parts. The position and confidence of the overall prediction boxes are calculated from the detected overall features of the targets. However, focusing only on the overall features of the target will often lose a lot of useful information, so we should pay attention to the more simple and obvious local features. More effective information can be obtained from the local feature prediction box, and the position of the overall prediction box can be restored. This method can effectively prevent missing detection in the case of occlusion, and the cooperation detection between different local features can achieve better detection results.

\section{Methodologies}\label{sec3}

In this chapter, we start by introducing the YOLOv5 model and the overall framework of the improved YOLOv5s with feature fusion module (FFM). After that, we described our improvements to the YOLOv5 network. Finally, we analyze the composition and working principle of our feature fusion module (FFM) in detail. To make the network much stronger, we redesign the loss function of YOLOv5.

\begin{figure*}[h]%
	\centering
	\includegraphics[width=0.9\textwidth]{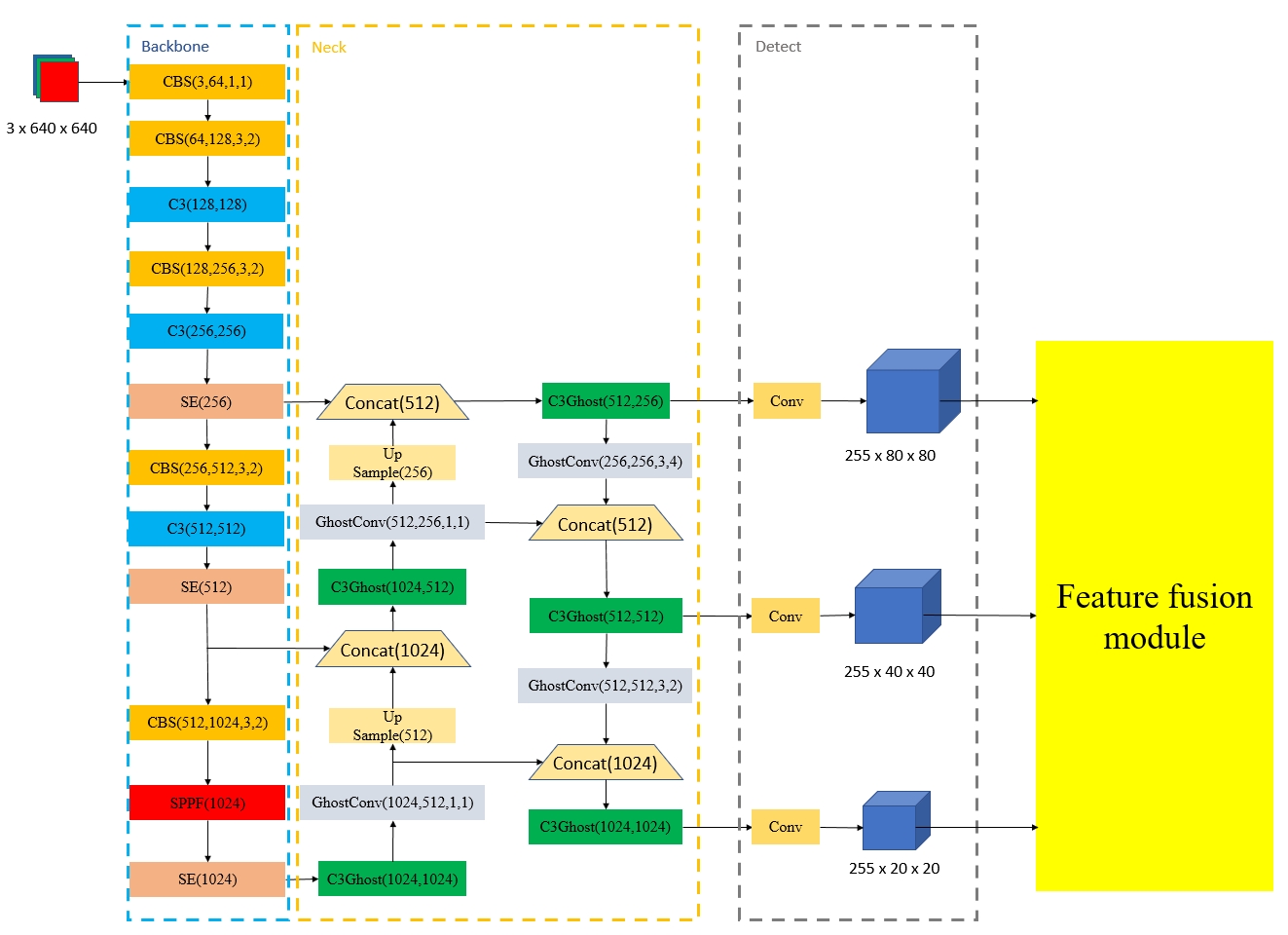}
	\caption{The structural diagram of the improved YOLOv5s with FFM}\label{fig1}
\end{figure*}

\subsection{YOLOv5 model}\label{subsec2}

According to the complexity of the model and the number of convolutions, YOLOv5 can be roughly divided into the following four mainstream versions: YOLOv5s, YOLOv5m, YOLOv5l, and YOLOv5x. These models have a similar framework. YOLOv5s has the minimum number of parameters and FLOPs. YOLOv5x achieves higher detection accuracy with the most convolution operations, but the amount of computation it entails is unacceptable. In this paper, the lightweight of the model is one of our purposes, so we chose to improve on the original YOLOv5s model.

\subsection{Overview of the proposed network}\label{subsec7}

In consideration of global information acquisition and lightweight, we designed the structural diagram of the improved YOLOv5s with feature fusion module (FFM) as shown in Fig.~\ref{fig1}. First, we added the SE attention blocks (light orange color) to the backbone network of YOLOv5s, which takes advantage of global information by giving weight to each channel. Secondly, according to the preliminary test information, we only improved the convolution operation in the neck network of YOLOv5s and introduced Ghost convolution module. The improved CBS and C3 modules were called GhostConv (light gray color) and C3Ghost (green color) modules respectively. In order to deal with the occlusion problem, we creatively designed an FFM module (yellow color) and added it to the end of the YOLOv5s model. For ease of understanding, we write the number of input and output channels, the size of convolution kernels, and the size of step sizes after each module.

\subsection{Redesign of the YOLOv5s network architecture}\label{subsec11}

\subsubsection{Improvement of the backbone network}\label{subsubsec8}

\begin{figure*}[h]%
	\centering
	\includegraphics[width=0.9\textwidth]{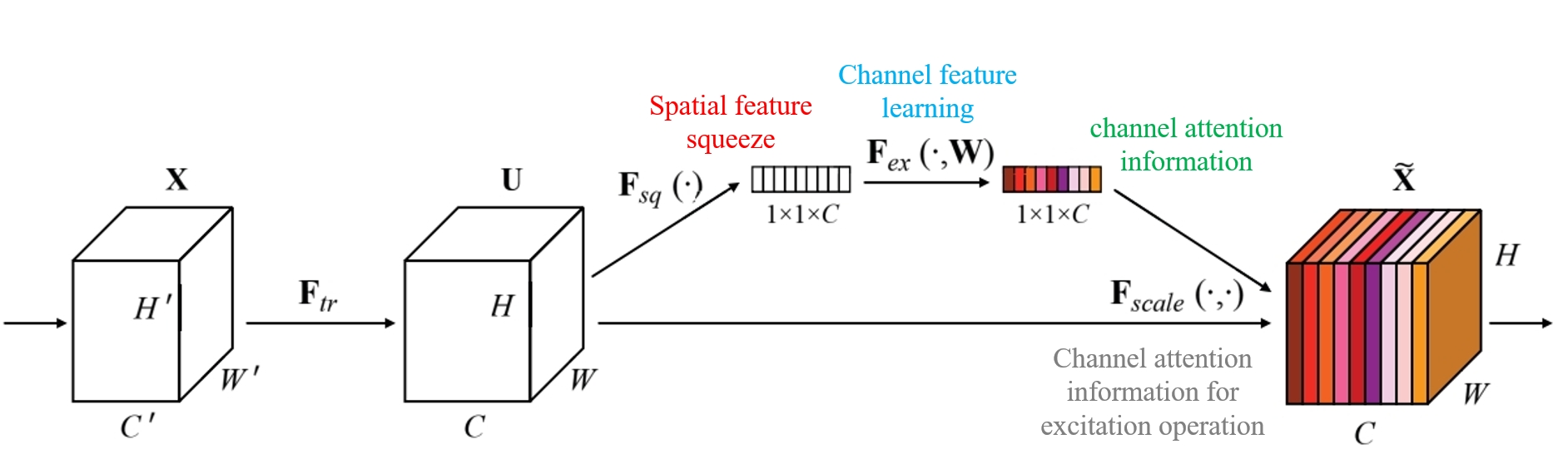}
	\caption{A squeeze-and-excitation block}\label{fig2}
\end{figure*}

In order to enhance the utilization of global information, the Squeeze-and-Excitation (SE) channel attention blocks have been added to the backbone network of YOLOv5s. The change of feature map in the SE block is shown in Fig.~\ref{fig2}. First, the SE block maps the input $X\in\mathbb{R}^{H^{\prime}{\times}W^{\prime}{\times}C^{\prime}}$ to $U\in\mathbb{R}^{H{\times}W{\times}C}$ via a transformation $F_{tr}$. $F_{tr}$ is a convolution operator and the learned set of filter kernels is represented by  $M=\left[m_1,m_2,...,m_c\right]$, where $m_1,m_2,...,m_c$ refers to the parameters of filter. $U=\left[u_1,u_2,...,u_c\right]$ is used to represent the output result after mapping:
\begin{equation}
	u_c = m_c * X = \sum\limits_{s=1}^{c^{\prime}}m^s_c * x^s.\label{eq1}
\end{equation}
Here $*$ represents the convolution operation, $m_c=\left[m^1_c,m^2_c,...,m^{c^{\prime}}_c\right]$, $X=\left[x^1,x^2,...,x^{c^{\prime}}\right]$ and $u_c\in\mathbb{R}^{H{\times}W}$. All channels of $X$ correspond to the parameter $m^s_c$ in $m_c$, which is a two-dimensional spatial kernel.

After that, the feature map performs a global average pooling operation on the spatial dimension through another transformation $F_{sq}$, and compresses the dimension from $H{\times}W{\times}C$ to $1{\times}1{\times}C$. Later, through the learning of the Fully Connected (FC) layer, the original feature map becomes a feature map with channel attention, and its dimension remains unchanged in this process. Finally, we multiply the original feature map with dimension $H{\times}W{\times}C$ and the obtained feature map with dimension $1{\times}1{\times}C$ channel by channel. This excitation operation restores the feature map of dimension $H{\times}W{\times}C$ and gives it channel attention.

\subsubsection{Improvement of the neck network}\label{subsubsec9}

The parameter number and FLOPs of the model are mostly determined by convolution operations. In order to reduce the complexity of the model, we abandoned some ordinary convolution operations in the YOLOv5s neck network and replaced them with Ghost modules \citep{bib11}. Fig.~\ref{fig3} and Fig.~\ref{fig4} compare the differences between the two convolution methods, and the specific calculation process is as follows:

\begin{figure}[h]%
	\centering
	\includegraphics[width=0.45\textwidth]{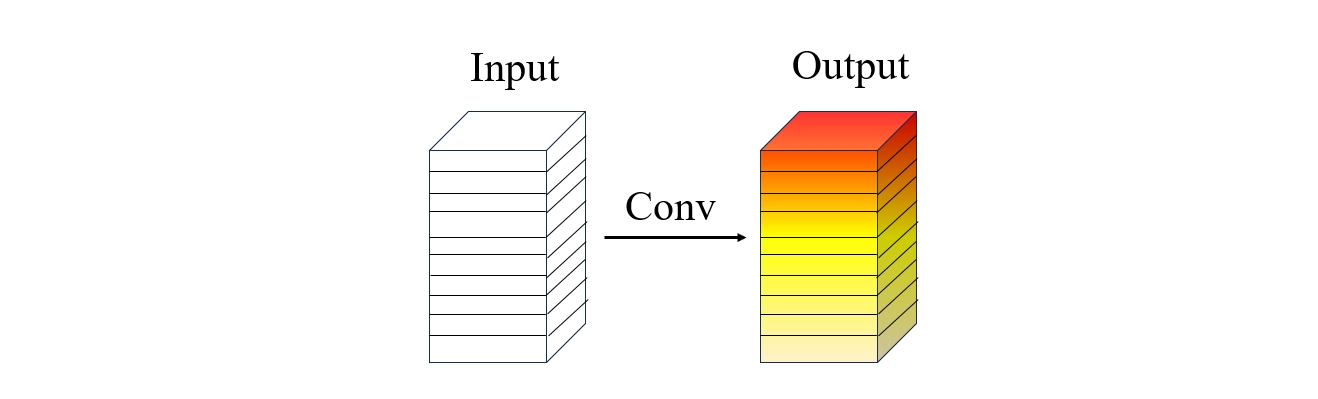}
	\caption{The ordinary convolution operations}\label{fig3}
\end{figure}

\begin{figure}[h]%
	\centering
	\includegraphics[width=0.45\textwidth]{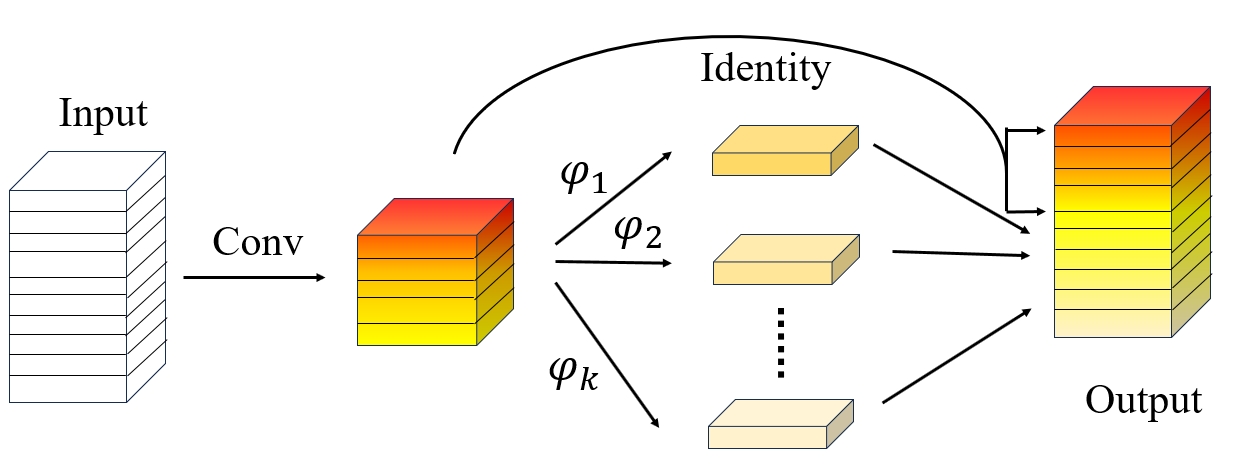}
	\caption{The Ghost module}\label{fig4}
\end{figure}

Suppose we use $h_1{\times}w_1{\times}c_1$ and $h_2{\times}w_2{\times}c_2$ to represent the size of the input and output feature maps, respectively. $h_1$ and $h_2$ represent the height of the input and output feature maps, $w_1$ and $w_2$ represent the width of the input and output feature maps. The size of the convolution kernel is denoted by $n{\times}n$. The required floating-point operations (FLOPs) to perform one convolution is $h_2{\times}w_2{\times}c_2{\times}c_1{\times}n{\times}n$, and this order of magnitude is usually greater than $10^5$.

Ghost module divides the convolutional layer into two parts during calculation, and adopts different calculation strategies for different parts. The first part is a conventional convolution, but the number of feature maps will be strictly controlled, because the calculation amount cannot be too large. Then the second part would also have to generate some feature maps, but instead of using conventional convolution, it would be generated by a simple "Linear Transformation". From the above description, we calculate the FLOPs of the Ghost module as $h_2{\times}w_2{\times}\frac{c_2}{s}{\times}c_1{\times}n{\times}n + (s-1){\times}h_2{\times}w_2{\times}\frac{c_2}{s}{\times}l{\times}l$, where $s$ represents the number of transformation operations performed at a low cost, and $s{\ll}c_1$, $l{\times}l$ refers to the kernel size of linear computation.

\begin{figure*}[h]%
	\centering
	\includegraphics[width=0.9\textwidth]{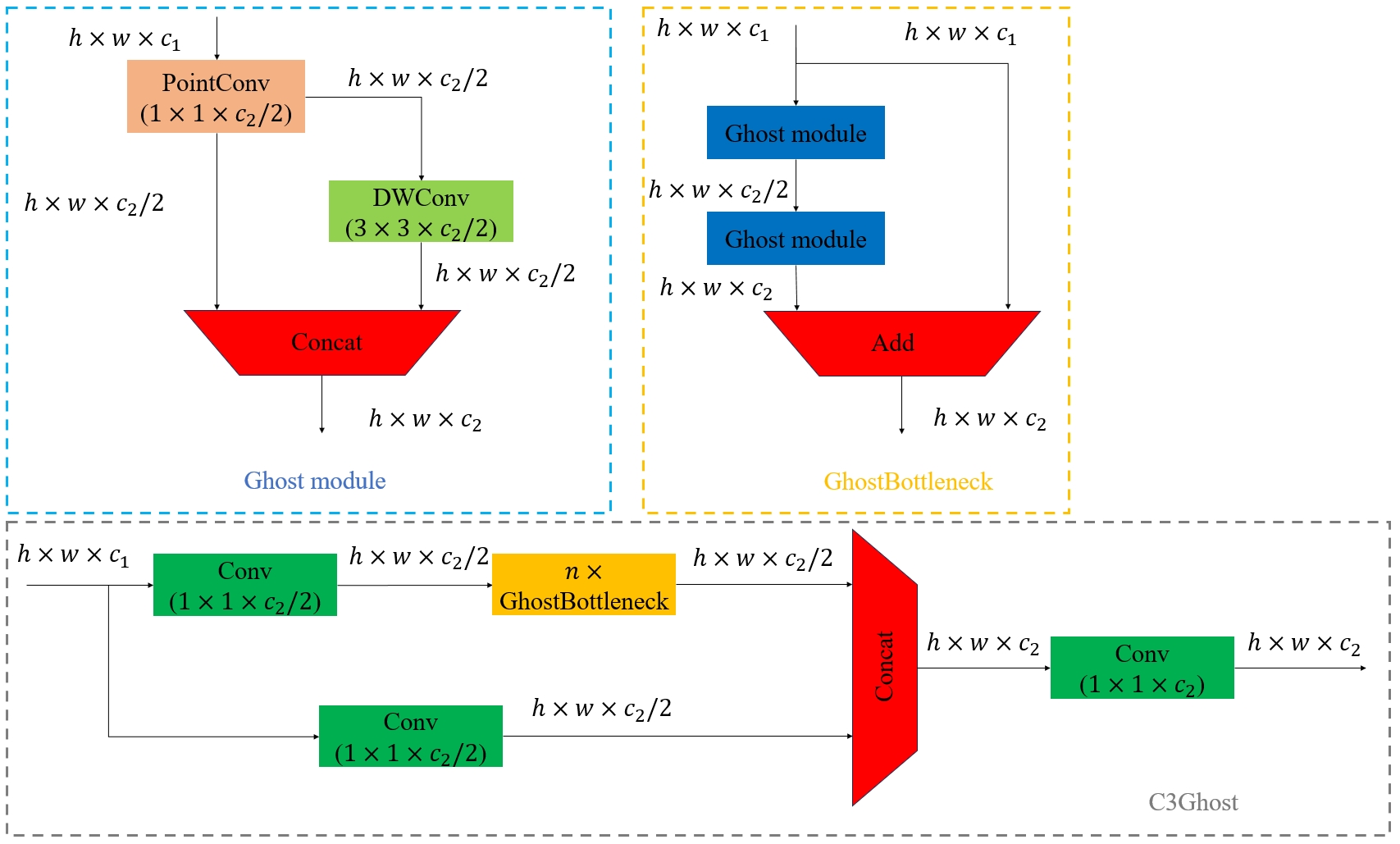}
	\caption{GhostBottleneck and C3Ghost}\label{fig5}
\end{figure*}

To quantify the improvements made by the Ghost module, we calculated the ratio of FLOPs required for both:
\begin{align}
	 r &= \frac{h_2{\times}w_2{\times}c_2{\times}c_1{\times}n{\times}n}{h_2{\times}w_2{\times}\frac{c_2}{s}{\times}c_1{\times}n{\times}n + (s-1){\times}h_2{\times}w_2{\times}\frac{c_2}{s}{\times}l{\times}l}  \nonumber \\
	 &=\frac{c_1{\times}n{\times}n}{\frac{1}{s}{\times}c_1{\times}n{\times}n+\frac{s-1}{s}{\times}l{\times}l} \approx  \frac{s{\times}c_1}{s+c_1-1} \approx s.\label{eq2}
\end{align}

According to formula \ref{eq2}, the FLOPs of the Ghost module are approximately $s$ times lower compared to the ordinary convolution. Based on the excellent performance of the Ghost module, we replace the standard convolution of the existing bottleneck and introduce the GhostBottleneck and C3Ghost modules, whose structures are shown in Fig.~\ref{fig5}. The new structure greatly reduces the computation and model complexity.

\subsubsection{Loss function improvement}\label{subsubsec7}

Because training data inevitably contain extreme samples. If the loss function still considers geometric measures such as distance and horizontal-to-vertical ratio, this will increase the penalty for extreme samples and lead to the deterioration of model generalization performance. In contrast, Wise-IoU \citep{bib12} weakens the penalty of geometric factors and focuses more on normal quality samples, so it has better generalization ability.

\subsection{Feature fusion module(FFM)}\label{subsec6}

\begin{figure*}[h]%
	\centering
	\includegraphics[width=0.9\textwidth]{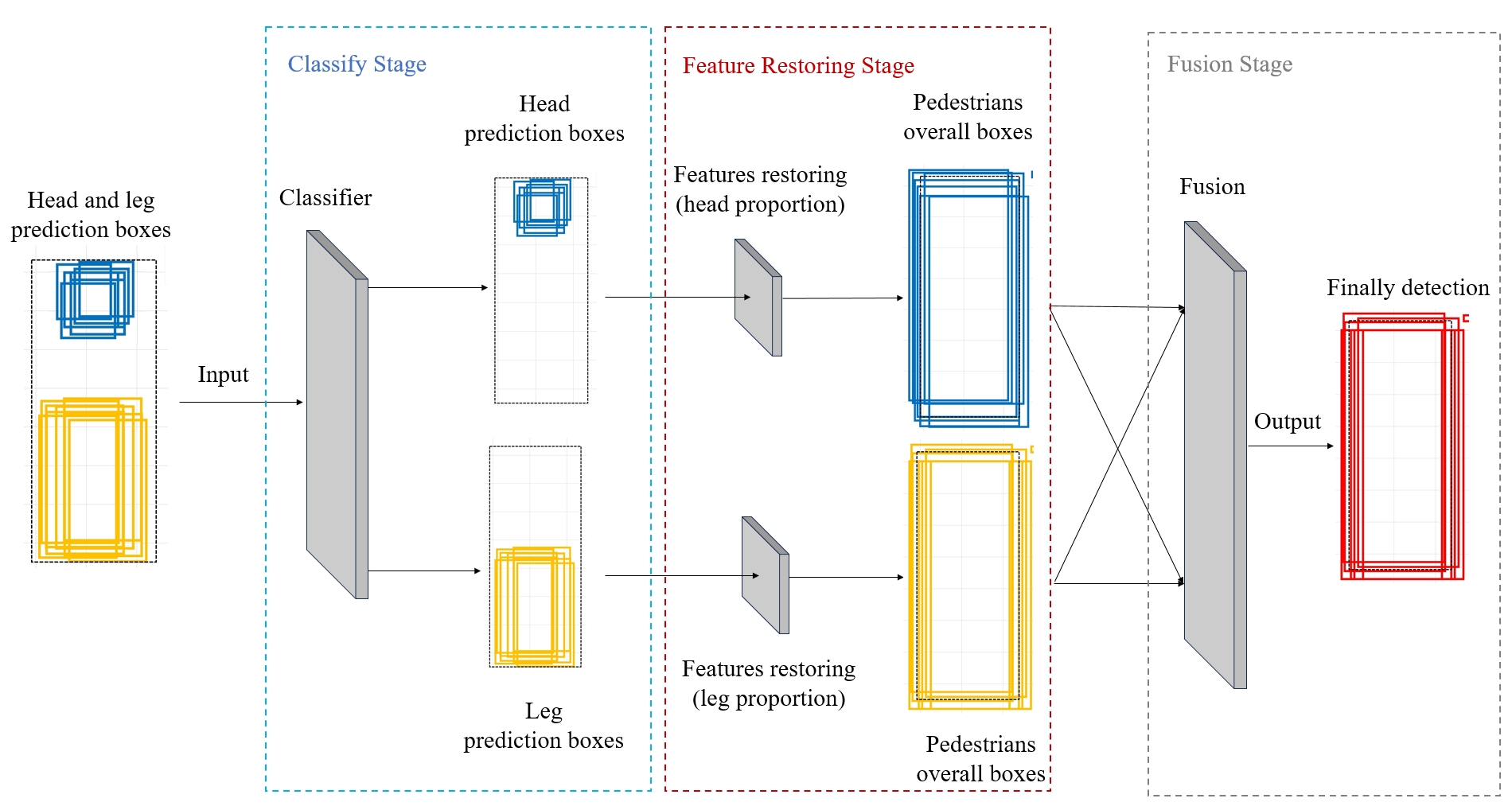}
	\caption{Feature fusion module(take the fusion of pedestrian head and leg features as an example)}\label{fig6}
\end{figure*}

In order to finally get the overall box of the pedestrian and deal with the occlusion problem, we propose a feature fusion module(FFM) to fuse the head and leg features of pedestrians, and its main architecture is shown in Fig.~\ref{fig6}. Our FFM consists of three stages as follows.

\textit{Classify stage}. In the first stage, the classifier divides the prediction boxes generated by YOLOv5 into $n$ sets according to different feature categories, Fig.~\ref{fig6} takes the detection of the head and leg area of pedestrians as an example, the blue bounding boxes represent the detected head region, the yellow bounding boxes represent the detected leg region. The classifier outputs two sets of different features. Each set contains all prediction boxes of the same type of features. Each prediction box consists of five types of information: class name, center point coordinates, height and width of prediction box, and confidence. There are two main reasons for putting different types of features into different sets. One is to avoid the fusion of restore boxes of similar classes, and the other is that the calculation only needs to be performed between the restore boxes of different classes instead of between the restore boxes of similar classes, which greatly reduces the amount of calculation.

\begin{figure}[h]%
	\centering
	\includegraphics[height=0.35\textheight]{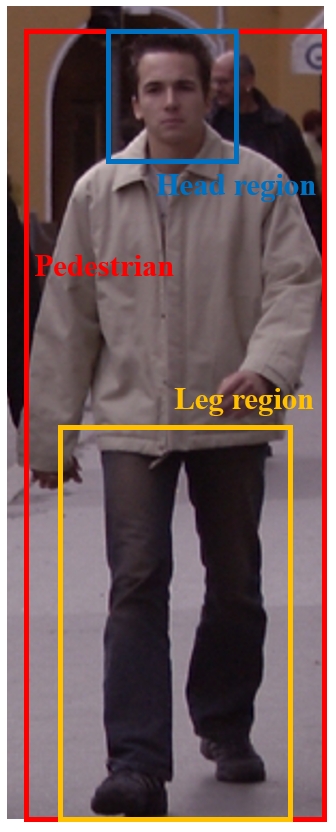}
	\caption{The inherent proportion of the human body}\label{fig7}
\end{figure}

\textit{Feature restoring stage}. In the second stage, we restore the local features to the overall prediction boxes by their fixed relationship. In this detection, we restore the head and leg regions of the pedestrian to the overall box of the pedestrian according to the inherent proportion of the human body (formula \ref{eq3}, Fig.~\ref{fig7}), respectively. 
\begin{align}
	&x_c = x_{c head}  ,  y_c = y_{c head} + 2H_{head} \nonumber \\
	&W = 2W_{head}  ,  H = 5H_{head} \nonumber \\
	&x_c = x_{c leg}  ,  y_c = y_{c leg} - \frac{1}{2}H_{leg} \nonumber \\
	&W = \frac{4}{3}W_{leg}  ,  H = 2H_{leg} \label{eq3}
\end{align}
where $x_{c head},y_{c head},x_{c leg},y_{c leg}$ represents the center point coordinates of the head region and leg region, $H_{head},W_{head},H_{leg},W_{leg}$ represents the height and width of the head area and the leg area, $x_c,y_c,H,W$ represents the coordinates of the pedestrian center point and the height and width of the pedestrian, and the class name and confidence remain the same. In this way, the prediction boxes of both sets are converted into pedestrian overall prediction boxes, but they come from different sources. In Fig.~\ref{fig6} feature restoring stage, the blue bounding boxes represent the overall boxes restored from the head area, while the yellow bounding boxes represent the overall boxes restored from the leg area.

\textit{Fusion stage}. In the third stage, we fuse these overall boxes restored by different local features. The IoU value will be calculated between each pedestrian overall box restored from the head area and all the pedestrian overall boxes restored from the leg area. The two types of restore boxes with the largest IoU and greater than the IoU threshold will be considered as the same pedestrian, and the two restore boxes with lower confidence will be deleted at the same time. Restoration boxes that do not match will be considered as occluded pedestrians. The remaining pedestrian overall boxes in the two sets are the final detection result (the red bounding boxes).

After decomposing the overall features into different local features, we improve the detection accuracy and generalization performance. However, for pedestrians without occlusion, we lose the internal relationship between the local features, which easily leads to the same pedestrian being restored by different features multiple times, resulting in repeated recognition. We connect these features through the Fusion stage, determine which features belong to the same person, and handle the occlusion problem with incomplete overall features well. 

\section{Experiments}\label{sec4}

We selected two commonly used pedestrian datasets with occluded situations, Citypersons dataset \citep{bib31} and CUHK Occlusion dataset \citep{bib32}, to conduct comparative experiments. Both quantitative and qualitative comparisons have been provided between improved YOLOv5s with FFM and other models in this series.

\textit{Implementation details}. We used Pytorch 2.1.2 with CUDA 12.0 as deep learning environments and conducted experiments on Ubuntu 20.04 systems. All models are tested on 1 NVIDIA GeForce RTX 3060 Ti video card with 8 GB memory.

\begin{figure}[h]%
	\centering
	\includegraphics[width=0.45\textwidth]{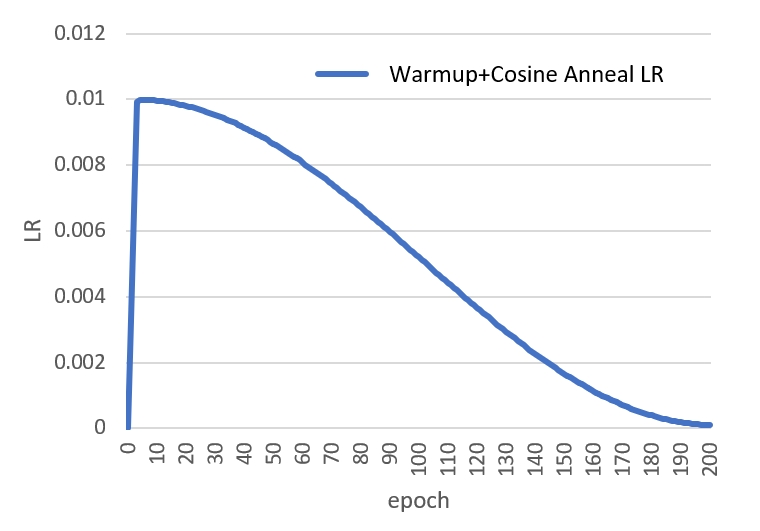}
	\caption{Warmup+Cosine Anneal LR}\label{fig8}
\end{figure}

\textit{Training details}. Learning rate is an important hyperparameter in YOLO training, which determines the step size of the model in each parameter update. At the beginning of training, we want to have a small learning rate to prevent the model from oscillating. As the model becomes stable, we need a large learning rate to get the convergence result as soon as possible. In the process of convergence, the learning rate needs to be continuously reduced. In order to meet the requirement of learning rate at each stage. We adopted the Warm-up \citep{bib34} learning rate adjustment strategy in all experiments, that is, at the beginning of training, we gradually increased the learning rate to a relatively appropriate value, which is set to 0.01, and then gradually reduced the learning rate according to the cosine annealing \citep{bib35} attenuation strategy. The learning rate for each training epoch is shown in Fig.~\ref{fig8}.

\textit{Evaluation metrics}. We use average precision (AP) to quantitatively evaluate our improved model and other existing models, the larger the result is, the better the performance is. In order to compare the complexity of the models, we further counted the number of parameters and FLOPs of each model. The smaller the two indicators, the lighter the model.

\subsection{Ablation study}\label{subsec9}

In order to compare the contribution of each element in our proposed model, Table~\ref{tab1} shows the results of different combinations of elements on Citypersons dataset \citep{bib31}.

\textit{SE block}. With SE block, the network can learn global information and assign weights to channels. Compared with YOLOv5s baseline, the AP has been improved by 1.3\% while the parameters and FLOPs were almost unchanged.

\textit{Ghost module}. With Ghost module, the network only needs to calculate a much smaller number of parameters to achieve almost the same effect as YOLOv5s baseline. As shown in Table~\ref{tab1}, the parameters went down by 28.8\% and the GFLOPs decreased by 19.6\%. In spite of these, only 0.2\% of the AP has been sacrificed.

\begin{table*}[h]
	\caption{Effects of various designs on Citypersons dataset}\label{tab1}%
	\begin{tabular}{@{}llllllll@{}}
		\toprule
		YOLOv5s baseline & SE block  & Ghost module & WIoU loss & FFM & AP(\%) & Parameters & GFLOPs\\
		\midrule
		\checkmark    &    &   &    &    &61.5  &7012822 &15.8 \\
		\checkmark    & \checkmark   &   &    &    &62.8 &7071158 &15.8  \\
		\checkmark    &    & \checkmark  &    &    &61.3 &4991654 &12.7  \\
		\checkmark    & \checkmark   & \checkmark  & \checkmark   &    &61.8 &5049990 &12.8  \\
		\checkmark    & \checkmark   & \checkmark  & \checkmark   & \checkmark   &$\mathbf{67.1}$ &$\mathbf{5052687}$ &$\mathbf{12.8}$  \\
		\botrule
	\end{tabular}
\end{table*}

\textit{WIoU loss}. The new loss function weakens the penalty caused by extreme samples and makes the model pay more attention to normal samples.

\textit{Feature fusion module}. In order to deal with the widespread occlusion problems in pedestrian dataset, we first used the improved YOLOv5s model to obtain the visible local features of pedestrians, and they are further processed by the feature fusion module (FFM). Compared with the YOLOv5s baseline, the AP has been boosted from 61.5\% to 67.1\%. This result was achieved while significantly reducing the number of parameters and FLOPs.

The above results prove that the improved YOLOv5s with FFM can achieve better average precision (AP) with fewer parameters and FLOPs.

\subsection{Experiment on Citypersons dataset}\label{subsec8}

The CityPersons dataset \citep{bib31} is a subset of the cityscape that consists only of person annotations. It consists of 18 training sets and 3 validation sets, which has a total of more than 3000 images. We used the warm-up strategy in the first 3 epochs of training the Citypersons dataset to ensure that the learning rate was not too high, and the cosine annealing strategy was used in the subsequent epochs to gradually reduce the learning rate.

In order to quantitatively illustrate the superiority of our proposed model, we trained all YOLOv5 versions, which have similar frameworks but different network depth and feature map width. However, they all achieve larger AP by adopting more parameter counts and GFLOPs. The experimental results of improved YOLOv5s with FFM and other models are shown in Table~\ref{tab2}. By adding FFM and improving the structures of YOLOv5s, our model achieves a higher AP on pedestrian detection. To be specific, the proposed model improves the AP by 5.6\%, reduces the parameters by 27.9\% and reduces the GFLOPs by 19.0\% compared with original YOLOv5s. Our model maintains an accuracy close to YOLOv5x, but the number of parameters and GFLOPs are reduced by 94.1\% and 93.7\% respectively.

\begin{table*}[h]
	\caption{Results of different models on the Citypersons dataset}\label{tab2}%
	\begin{tabular}{@{}llllll@{}}
		\toprule
		Model & YOLOv5s  & YOLOv5m & YOLOv5l & YOLOv5x & ours(Improved YOLOv5s with FFM)\\
		\midrule
		Input    & 640 x 640   & 640 x 640  & 640 x 640  & 640 x 640  & 640 x 640  \\
		Parameters    & 7012822   & 20852934  &46108278   &86173414   & $\mathbf{5052687}$  \\
		GFLOPs    & 15.8   & 47.9  &107.6    &203.8    & $\mathbf{12.8}$  \\
		AP(\%)    & 61.5   & 64.8  &66.4    &67.6    & $\mathbf{67.1}$  \\
		\botrule
	\end{tabular}
\end{table*}

\begin{figure*}[h]%
	\centering
	\includegraphics[height=0.67\textheight]{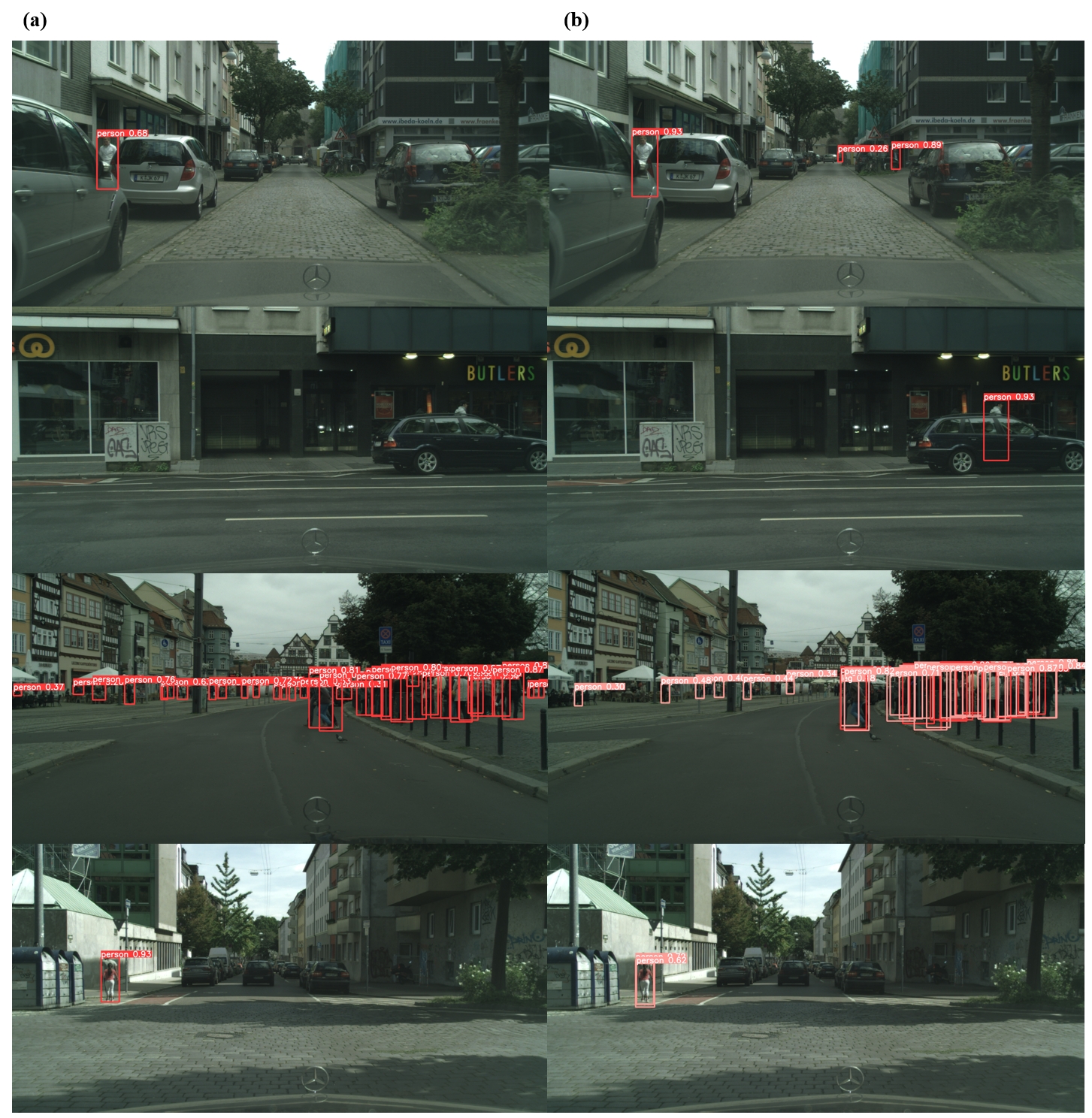}
	\caption{Comparative test results of YOLOv5s and improved YOLOv5s with FFM on Citypersons dataset. $\mathbf{(a)}$ YOLOv5s $\mathbf{(b)}$ Improved YOLOv5s with FFM}\label{fig9}
\end{figure*}

Fig.~\ref{fig9} shows the detection results of YOLOv5s and improved YOLOv5s with FFM on Citypersons dataset. Figure ~\ref{fig9}a is the result of detection by original YOLOv5s, indicating that the effect of interclass occlusion or intra-class occlusion is not good. Fig.~\ref{fig9}b is the result of detection by improved YOLOv5s with FFM, as we can see, our method has handled these cases well.

\subsection{Experiment on CUHK Occlusion dataset}\label{subsec10}

We also run the proposed model on CUHK Occlusion dataset \citep{bib32} for comparative experiments. In addition to pedestrians on the road, CUHK Occlusion dataset also includes some pedestrians on campus and from the perspective of indoor surveillance, which has a much richer scene. There are 9 scenarios in the dataset, and we trained all the models with 50 epochs on each of these scenarios separately. Similarly, we used the warm-up strategy in the first 3 training epochs and updated the learning rate with the cosine annealing algorithm in the subsequent process.

Table~\ref{tab3} displays the average precision (AP) of the different models for each scenario of the CUHK Occlusion dataset. It is clear that the improved YOLOv5s with FFM performs better than YOLOv5s original models in all 9 scenarios on CUHK Occlusion dataset. In some scenarios, the average precision of our model is even higher than YOLOv5x, which requires calculating 17 times more parameters than our model.

\begin{table*}[h]
	\caption{Result of different models on CUHK Occlusion dataset(each Seq represents a different scene)}\label{tab3}%
	\begin{tabular}{@{}llllll@{}}
		\toprule
		Model & YOLOv5s  & YOLOv5m & YOLOv5l & YOLOv5x & ours(Improved YOLOv5s with FFM)\\
		\midrule
		Seq 0    & 80.4\%   & 82.9\%  & 83.5\%  & 84.6\%  & $\mathbf{89.6\%}$  \\
		Seq 1    & 82.2\%   & 84.0\%  & 87.9\%  & 88.5\%  & $\mathbf{85.6\%}$  \\
		Seq 2    & 87.9\%   & 92.2\%  & 93.3\%  & 94.8\%  & $\mathbf{89.3\%}$  \\
		Seq 3    & 94.8\%   & 96.2\%  & 97.0\%  & 97.3\%  & $\mathbf{98.1\%}$  \\
		Seq 4    & 87.9\%   & 88.7\%  & 91.3\%  & 91.6\%  & $\mathbf{96.1\%}$  \\
		Seq 5    & 79.9\%   & 88.4\%  & 89.6\%  & 90.6\%  & $\mathbf{93.7\%}$  \\
		Seq 6    & 99.2\%   & 99.3\%  & 99.4\%  & 99.4\%  & $\mathbf{99.5\%}$  \\
		Seq 7    & 83.8\%   & 88.0\%  & 89.8\%  & 90.1\%  & $\mathbf{89.1\%}$  \\
		Seq 8    & 98.8\%   & 99.3\%  & 99.4\%  & 99.4\%  & $\mathbf{99.2\%}$  \\
		\botrule
	\end{tabular}
\end{table*}

\begin{figure*}[h]%
	\centering
	\includegraphics[width=0.95\textwidth]{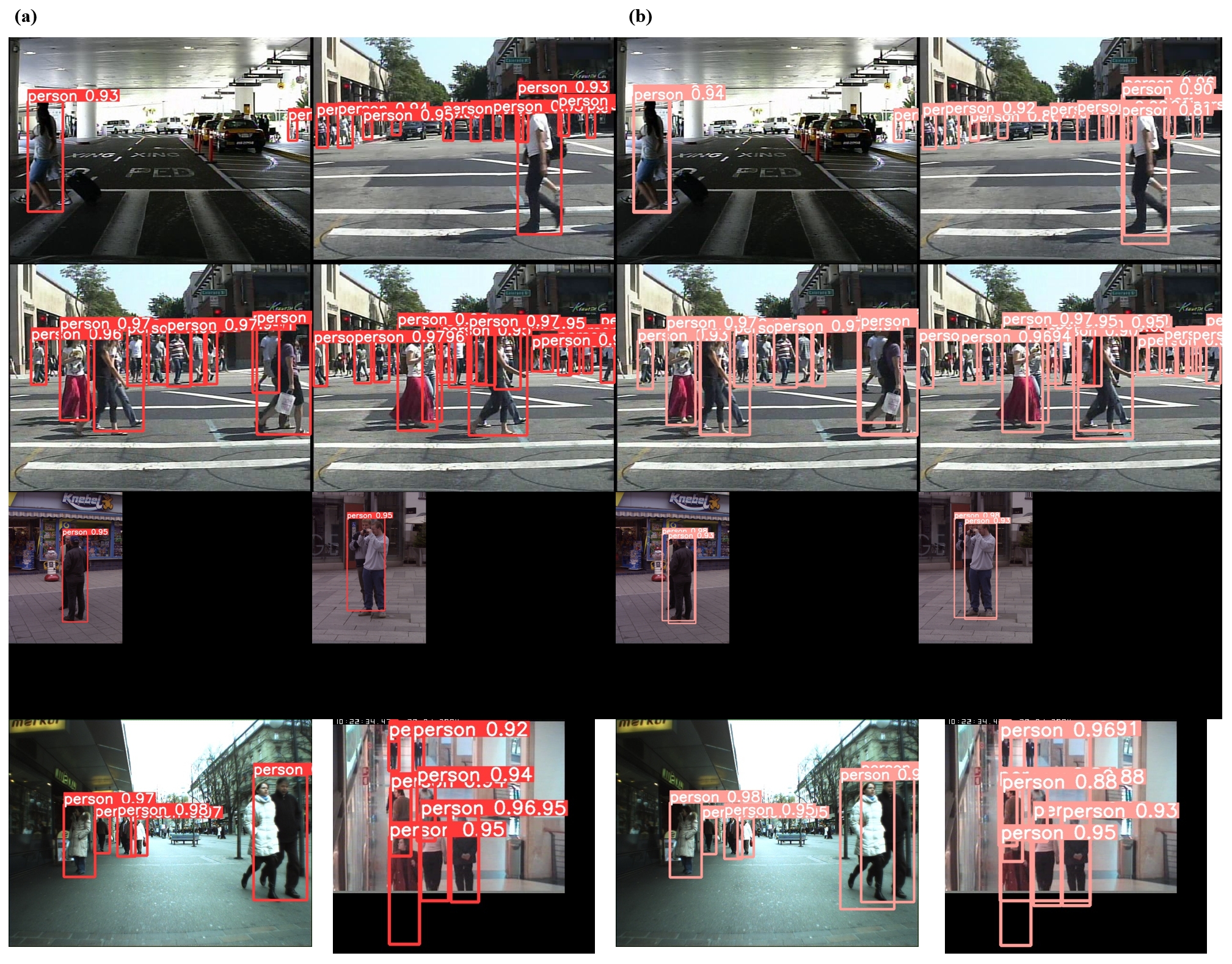}
	\caption{Comparative test results of YOLOv5s and improved YOLOv5s with FFM on CUHK Occlusion dataset. $\mathbf{(a)}$ YOLOv5s $\mathbf{(b)}$ Improved YOLOv5s with FFM}\label{fig10}
\end{figure*}

We show the different detection results of YOLOv5s baseline and improved YOLOv5s with FFM in Fig.~\ref{fig10}. As a qualitative comparison, Figure ~\ref{fig10}a is the result of detection by original YOLOv5s, Fig.~\ref{fig10}b is the result of detection by improved YOLOv5s with FFM. It is still not difficult to see the advancement of our proposed model in detecting complex scenes.

\subsection{Discussions}\label{subsec12}

The proposed model can significantly improve the average precision (AP) of pedestrian detection especially in the case of occlusion, that is because we detect different local features. While the previous algorithms are trained and detected for the whole pedestrian feature, these algorithms are not very robust to the occlusion situation. When the pedestrian is occluded, the proposed algorithm can still detect the local features of the visible area, and then restore the overall box of the pedestrian. 

Because our method can restore pedestrians from multiple local areas, it is almost impossible for all local features of one person to be occluded at the same time. Therefore, compared with directly detecting the whole pedestrian, the detection accuracy will be improved and it is not susceptible to the influence of non-maximum suppression(NMS). In addition, for targets that are not occluded or with few occluded areas, we can restore the overall box of the pedestrian through multiple local feature fusion, and the accuracy will be further improved.

Although our approach achieves better results than original YOLOv5s, it still falls short in some cases. For example, when a pedestrian appears in a very strange pose in the image, our model will restore the different local features to multiple overall boxes. When the IOU between each overall box of the same person is too small, it will be considered to be different pedestrians, resulting in virtual detection.

\section{Conclusion}\label{sec13}

We creatively propose a detection strategy to restore the whole pedestrian box by detecting local features, and design a feature fusion module (FFM) for this strategy, which was successfully added to the YOLO detector. The main contributions are divided into the following points: First, we designed a feature fusion module (FFM) for occlusion detection, instead of simply increasing the complexity of the model to learn various occlusion situations. On the contrary, we replace traditional convolution with the Ghost module, reducing model parameters and FLOPs. In addition, we introduced SE blocks to the YOLOv5s backbone network and gave it the ability to consider global information. Finally, to establish an efficient detector that can handle occlusion problems, we combine improved YOLOv5s with FFM and use WIoU to redesign the loss function. We conducted quantitative and qualitative experiments on Citypersons dataset and CUHK Occlusion dataset. The detection data shows that the improved YOLOv5s with FFM enjoys better average precision (AP) and less parameter calculation than the existing excellent object detection models.

\backmatter

\bmhead{Acknowledgments}

We thank the anonymous editor and reviewers for their careful reading and many insightful comments and suggestions.

\bmhead{Author Contributions}

Conceptualization: X.L., B.S.; Methodology: X.L.; Formal analysis and investigation: X.L.; Writing-original draft preparation: X.L.; Writing-review and editing: X.L., B.S., Z.C., Y.W; Resources: X.L.; Supervision: Z.C., Y.W.

\bmhead{Funding} 

No funding was received to assist with the preparation of this manuscript.

\bmhead{Availability of data and materials}

The datasets used in our experiments are public datasets. Additional annotation files and test results are available from the author on reasonable request.

\section*{Declarations}

\bmhead{Competing interests}

The authors declare that they have no financial or proprietary interests in any material discussed in this article.

\bmhead{Ethics approval}

Not applicable.


\bibliography{sn-bibliography}


\end{document}